\PassOptionsToPackage{table,dvipsnames}{xcolor}
\documentclass[10pt,twocolumn,letterpaper]{article}
\usepackage[pagenumbers]{iccv} 

\definecolor{iccvblue}{rgb}{0.21,0.49,0.74}
\usepackage[pagebackref,breaklinks,colorlinks,allcolors=iccvblue]{hyperref}
\usepackage{multirow}
\usepackage{graphicx}
\usepackage{algorithm}
\usepackage{algpseudocode}
\usepackage{amsmath}
\def\paperID{6238} 
\def\confName{ICCV}
\def\confYear{2025}
\usepackage{amsmath}
\title{LIRA: Inferring Segmentation in Large Multi-modal Models with Local Interleaved Region Assistance}

\author{
Zhang Li\textsuperscript{1}, Biao Yang\textsuperscript{1}, Qiang Liu\textsuperscript{2}, Shuo Zhang\textsuperscript{1}, Zhiyin Ma\textsuperscript{1}, \\ 
Liang Yin\textsuperscript{1},
Linger Deng\textsuperscript{1}, 
 Yabo Sun\textsuperscript{2},
Yuliang Liu\textsuperscript{1}$^*$,
Xiang Bai\textsuperscript{1} \\ 
\\
\textsuperscript{1}Huazhong University of Science and Technology, \textsuperscript{2}Kingsoft Office\\
{\tt\small \{zhangli123, ylliu\}@hust.edu.cn}
}

\begin{document}

\maketitle
\begin{abstract}
    While large multi-modal models (LMMs) demonstrate promising capabilities in segmentation and comprehension, they still struggle with two limitations: inaccurate segmentation and hallucinated comprehension. These challenges stem primarily from constraints in weak visual comprehension and a lack of fine-grained perception. To alleviate these limitations, we propose LIRA, a framework that capitalizes on the complementary relationship between visual comprehension and segmentation via two key components: (1) Semantic-Enhanced Feature Extractor (SEFE) improves object attribute inference by fusing semantic and pixel-level features, leading to more accurate segmentation; (2) Interleaved Local Visual Coupling (ILVC) autoregressively generates local descriptions after extracting local features based on segmentation masks, offering fine-grained supervision to mitigate hallucinations. Furthermore, we find that the precision of object segmentation is positively correlated with the latent related semantics of the \texttt{<seg>} token. To quantify this relationship and the model's potential semantic inferring ability, we introduce the Attributes Evaluation (AttrEval) dataset. Our experiments show that LIRA achieves state-of-the-art performance in both segmentation and comprehension tasks. Code will be available at \url{https://github.com/echo840/LIRA}.

    
\end{abstract}

\let\thefootnote\relax\footnotetext{\textsuperscript{$^*$} Corresponding author}

\section{Introduction}
\label{sec:intro}

    With the development of large multi-modal models, their capabilities have expanded from visual comprehension to pixel-wise segmentation by integrating LMMs with segmentation modules. The pioneering work LISA~\cite{lisa} first introduced the embedding-as-mask paradigm to unlock segmentation capabilities and proposed the task of reasoning segmentation, which requires models to generate binary segmentation masks based on implicit text queries. Recently, OMG-LLaVA~\cite{omg-llava} used a universal segmentation method as a visual encoder, integrating image information and perception priors into visual tokens provided to large language models (LLM), thus better supporting both segmentation and comprehension.

    \begin{figure}[t]
      \centering
       \includegraphics[width=0.9\linewidth]{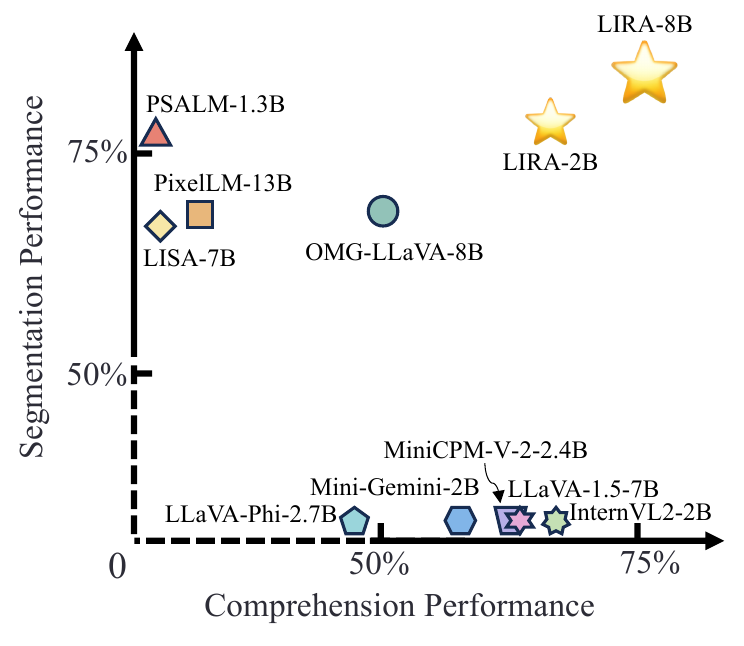}
       \caption{The performance of LIRA on comprehension and segmentation tasks compared with existing models. The x-axis represents the scores obtained by normalizing MME~\cite{mme} to a percentage scale. The y-axis represents the average performance on the RefCOCO~\cite{refcoco}, RefCOCO+~\cite{refcoco}, and RefCOCOg~\cite{refcocog} datasets. LIRA achieves excellent performance on both tasks.} 
       \label{fig:intro1}
    \end{figure}
    
\begin{figure}
  \centering
   \includegraphics[width=1\linewidth]{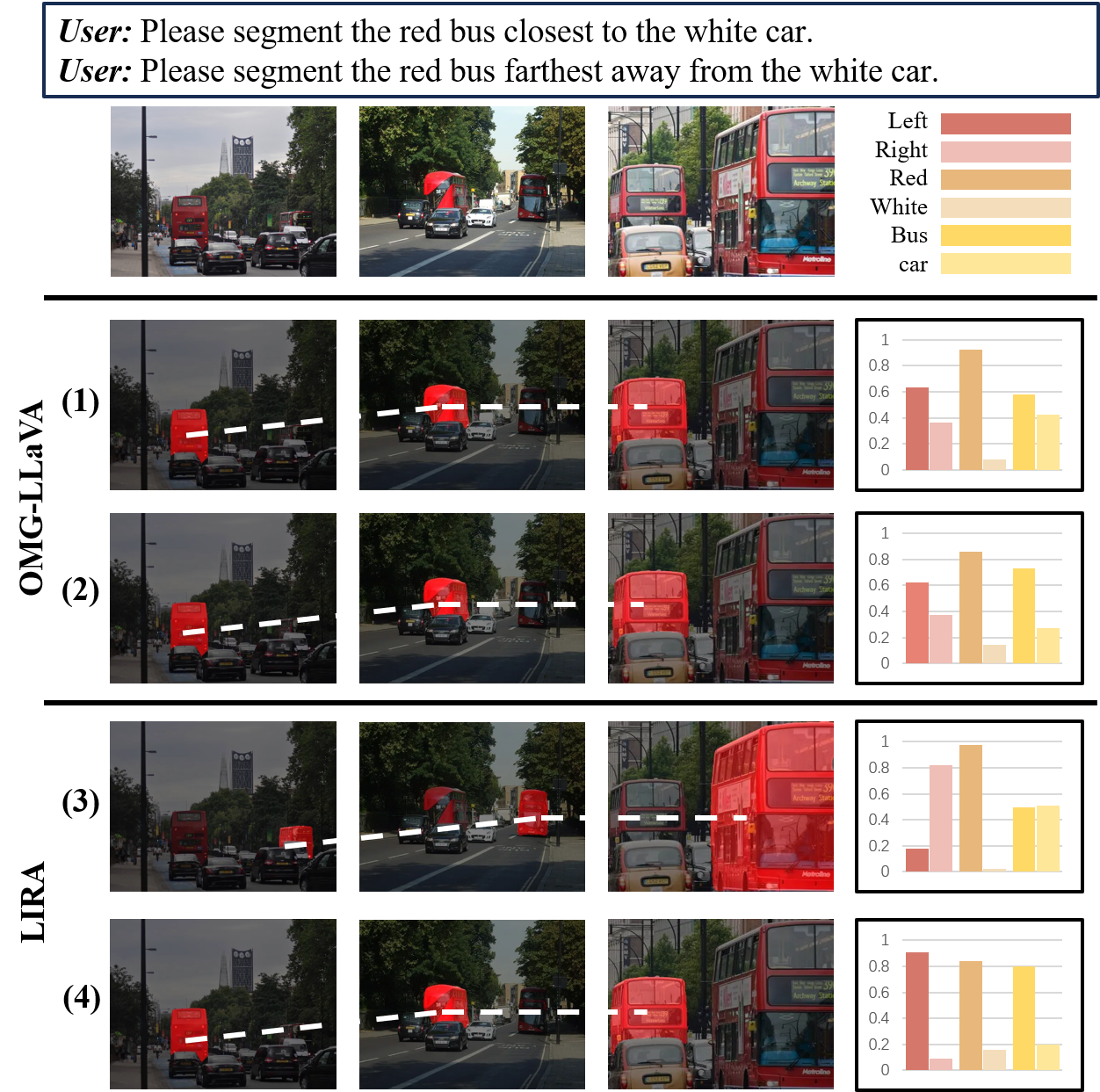}
   \caption{``Please segment the red bus closest to the white car." is used for the (1) and (3) rows of images in the first column, and ``Please segment the red bus farthest away from the white car." is used for the (2) and (4) rows of images in the first column. The dashed lines indicate that the three images share the \texttt{<seg>} token of the image in the first column. The right of the image represents the logits of the \texttt{<seg>} token of the images in the first column. 
   }
   \label{fig:analysis}
\end{figure}

    Despite significant advancements, previous methods often failed to accurately segment objects when meeting complex scenarios. As shown in Fig.~\ref{fig:analysis}, OMG-LLaVA failed to segment ``the red bus closest to the white car". To investigate the causes of segmentation errors, we extracted the embedding of the \texttt{<seg>} token from LMM using the image in the first column and applied it to segment the images in the second and third columns without further interaction with LLM. Interestingly, we observed that the bus on the left was consistently segmented in all images in row (1). Upon analyzing the logits of the \texttt{<seg>} token, we found that the higher logits for ``left" caused the bus on the left to be segmented. We suspect that the error occurred because the LMM failed to effectively embed accurate positional information into the \texttt{<seg>} token due to the limited visual comprehension of the LMM.
    
    Additionally, previous methods typically rely on position queries to indicate object locations and do not always establish a clear connection between local descriptions and the corresponding local image features, which may lead to hallucinations. This prompts an important question: Would it be more effective to directly input local image features into the LLM, enabling the model to generate region-specific descriptions and establish a more explicit mapping between visual features and their corresponding descriptions?

    In this paper, we propose LIRA, \textbf{L}ocal \textbf{I}nterleaved \textbf{R}egion \textbf{A}ssistance, incorporating two key components: 1) The Semantic-Enhanced Feature Extractor (SEFE) fuses high-level semantic comprehension with fine-grained pixel features through parametric fusion, enhancing visual comprehension for more precise object segmentation; 2) The Interleaved Local Visual Coupling (ILVC) enhances alignment between visual cues and textual descriptions by coupling image regions with corresponding text, reducing hallucination and enabling more accurate responses. As illustrated in Fig.~\ref{fig:analysis}, LIRA accurately interprets queries such as ``the red bus closest to the white car” as referring to the ``right bus", allowing for precise segmentation. 
    This process involves understanding object attributes based on both the query and the image to achieve accurate segmentation, which we refer to as inferring segmentation. This may differ from the definition used in LISA Reasoning Segmentation, which relies on external or commonsense knowledge to reason about implicit queries (e.g., “the food rich in Vitamin C”).
    As shown in Fig.~\ref{fig:intro1}, LIRA demonstrates strong performance across a range of standard comprehension and segmentation tasks. We summarize the main contributions as follows:
    \begin{itemize}
        \item \textbf{Improved Segmentation Accuracy}. LIRA incorporates the Semantic-Enhanced Feature Extractor (SEFE), which effectively integrates high-level semantic features with fine-grained pixel-level information. This fusion significantly enhances the model's ability to infer object attributes, resulting in more precise segmentation outcomes.

        \item \textbf{Reduced Hallucination}. Through the Interleaved Local Visual Coupling (ILVC) mechanism, LIRA establishes  a robust alignment between local image regions and their corresponding textual descriptions. 
        By explicitly coupling visual features with their semantic counterparts, LIRA reduces hallucinated comprehension and ensures the generation of precise and accurate image descriptions.

        \item \textbf{Strong Performance in Both Segmentation and Comprehension Tasks}. LIRA demonstrates strong performance on both segmentation and visual comprehension benchmarks. Specifically, experiments show that LIRA results in only a slight decrease compared to training with comprehension data alone (75.2\% vs. 75.3\%), significantly outperforms the previous best method, OMG-LLaVA, which shows a larger decline of (55.6\% vs. 69.9\%) across five comprehension datasets.
    \end{itemize}

\begin{figure*}[t]
  \centering
   \includegraphics[width=0.9\linewidth]{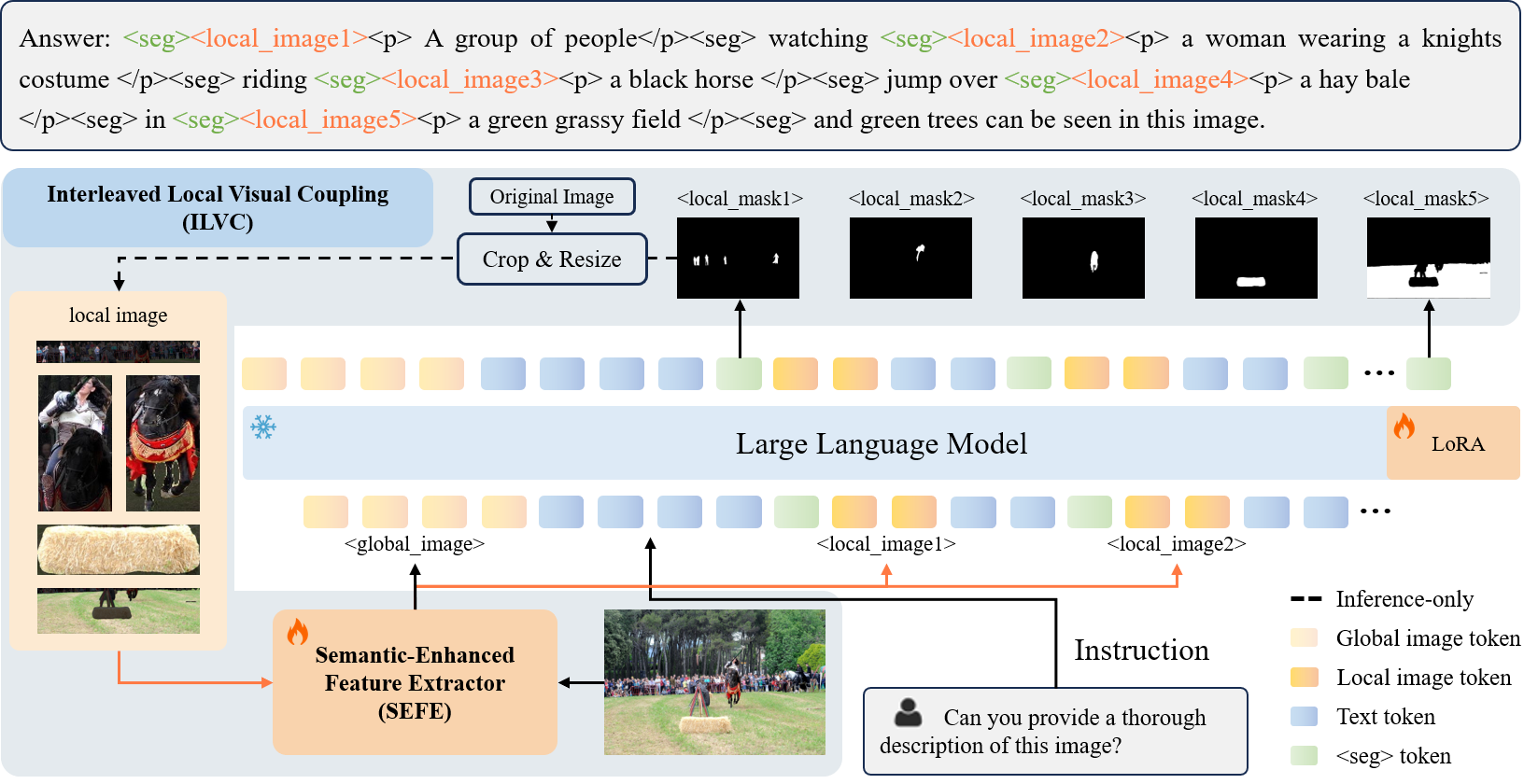}
   \caption{The overall architecture of LIRA. Global and local features are extracted using SEFE, with ILVC introduced to align \texttt{<mask, region, text>} correspondences. Note that in this pipeline, the \texttt{<seg>} token passing through a projector and pixel decoder for mask generation is omitted for simplicity. The orange line indicates the use of only the semantic encoder in SEFE, avoiding excessive zooming of local images in the pixel encoder.}
   \label{fig:arch}
\end{figure*}
\section{Related Work}
\label{sec:rela}
\subsection{Comprehension-Oriented LMMs}
\label{subsec:rela1}
    In recent years, the development of LMMs has made remarkable strides, improving the visual comprehension capabilities of LLMs. Early works~\cite{flamingo, blip2, llava, minigpt-4, mplug-owl} explore various methods to integrate visual features into Large Language Models (LLMs). Building on these foundations, subsequent approaches have focused on refining the quality of visual input, including techniques such as curriculum learning~\cite{qwen-vl, pali3}, high-resolution image cropping~\cite{ureader, monkey, llava-next, llava-uhd, textmonkey}, and separating high- and low-resolution branches~\cite{cogagent, llava-hr}. Additionally, advancements like chain-of-thought reasoning~\cite{vstar, cogcom, chain} and the mixture of experts framework~\cite{cogvlm, mplug-owl2} have further enhanced LMMs' ability to tackle complex visual comprehension tasks. More recently, works such as Cambrian~\cite{cambrian}, InternVL2~\cite{internvl2}, mPLUG-Owl3~\cite{mplug-owl3}, and Qwen2-VL~\cite{qwen2-vl} have made significant improvements in high-resolution image understanding and long-sequence visual input processing. InternLM-XComposer-2.5~\cite{internlm} has shown promising results in multi-turn dialogue and more efficient visual representations.

    \subsection{Segmentation-Oriented LMMs}
    \label{Subsec:rela1}
    The advancement of large multi-modal models (LMMs) has led to the integration of specialized task modules for vision-centric tasks. Some models~\cite{gpt4roi, shikra, kosmos2, ferret} generally focus on detection by expanding their functionality to support localization tasks through coordinate-based text outputs. 
    VisionLLM~\cite{visionllm} adds additional tokens designed for vision-centric tasks to the vocabulary of LLM, transforming the object localization task from continuous variable prediction to a more unified discrete bin classification.
    To facilitate pixel-level perception, other works~\cite{interngpt, llava-plus,sam4mllm,text4seg}, such as Text4Seg~\cite{text4seg}, employ LMMs as agents to collaborate with various visual models, such as SAM~\cite{sam}. Although these works are simple and effective, they cannot be end-to-end optimized.
    To harmonize model optimization, some approaches~\cite{lisa,lisa++,glamm,pixellm,psalm,visionllmv2}, leverage additional decoders to decode the \texttt{<seg>} tokens of the LMM. For instance, LISA~\cite{lisa} introduces an embedding-as-mask paradigm, utilizing special \texttt{<seg>} tokens to trigger segmentation pixel decoders like SAM~\cite{sam}. 
    PixelLM~\cite{pixellm} replaces SAM with a lightweight pixel decoder and introduces a comprehensive segmentation codebook to enhance efficient multi-object reasoning and segmentation.
    SegLLM~\cite{segllm} uses HIPIE-R50~\cite{hipie} as a pixel decoder and achieves novel interactive reasoning segmentation by reintegrating previous segmentation results into its input stream. 
    READ~\cite{qian2024reasoning} infers the semantics of the  \texttt{<seg>} token by analyzing its similarity to image features, whereas our method directly examines the logits of the  \texttt{<seg>} token in the language model.
    Recently, OMG-LLaVA~\cite{omg-llava} OMG-LLaVA proposes a method to integrate image information, perception priors, and visual prompts into visual tokens for language models (LLMs), but highlights that co-training with segmentation data can diminish comprehension performance in LLMs when applied to segmentation tasks. In this work, we discuss the complementarity of visual comprehension and segmentation, leveraging their interplay to improve segmentation accuracy and reduce hallucinations.

\section{Methodology}
\label{Sec:method}
\subsection{Overview}
    Fig.~\ref{fig:arch} illustrates the overall pipeline of LIRA, which comprises two key components: the Semantic-Enhanced Feature Extractor (SEFE) and the Interleaved Local Visual Coupling (ILVC) training paradigm.
    For the input image, SEFE first extracts the detailed features with dynamic cropping, obtaining a comprehensive representation of the image. The encoded image features are then combined with the instruction to be fed into the LLM.
    In the generating process, we introduce Interleaved Local Visual Coupling (ILVC) to ensure fine-grained alignment between image elements and their descriptions. We use the \texttt{<seg>} token to generate the segmentation mask, which is then used to extract the segmented image regions from the original image, resized to 448$\times$448, and input into SEFE to extract the local features. Subsequently, these encoded local features are re-input into the LLM to generate descriptions of the image regions and predict the subsequent content.

    \begin{figure}[t]
      \centering
       \includegraphics[width=0.95\linewidth]{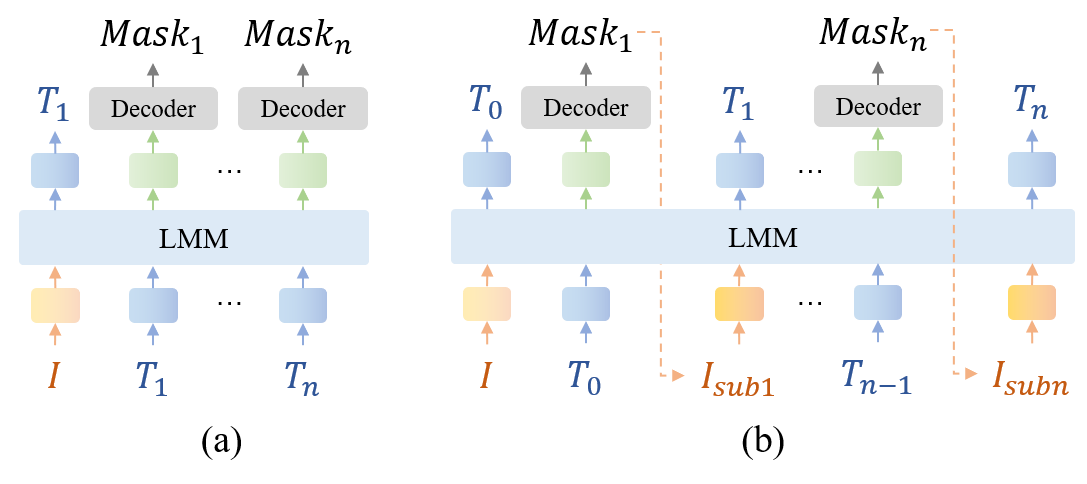}
       \caption{Different training paradigms: (a) refines only by refeeding the \texttt{<seg>} token into the model, while (b) extracts the corresponding image region based on the segmentation mask and refines by refeeding both the \texttt{<seg>} token and the local image region into the model. Refer to the color legend in Fig.~\ref{fig:arch} for details.}
       \label{fig:paradigm}
    \end{figure}
 
\subsection{Semantic-Enhanced Feature Extractor}
\label{method:sefe}
  
    To enhance segmentation performance, OMG-LLaVA~\cite{omg-llava} utilized a universal segmentation method as the visual encoder to integrate image information and segmentation priors into visual tokens; however, as shown in Fig.~\ref{fig:analysis}, OMG-LLaVA has lower logits value for the correct position, leading to inaccurate segmentation.
    To bridge the gap in visual comprehension and pixel reasoning, we introduce the Semantic-Enhanced Feature Extractor (SEFE), which organically integrates the advantages of a semantic encoder from a pre-trained LMM  and a pixel encoder sourced from a segmentation model.
    Given a global image $I \in \mathbb{R}^{3 \times H \times W}$, the semantic encoder and the pixel encoder extract visual features, respectively. The extracted features are then fed into MLP to transform the vision feature with the same dimension, which can be formulated as:
    \begin{equation}
    f_s = \mathrm{MLP}_s(\mathcal{S}(\mathbf{I})), \quad f_p = \mathrm{MLP}_p(\mathcal{P}(\mathbf{I})),
    \end{equation}
    where $\mathcal{S}$ and $\mathcal{P}$ represent the semantic encoder and the pixel encoder, respectively. 
    Next, we introduce a feature fusion module that leverages the semantic information from the semantic encoder to align the segmentation priors embedded in the pixel encoder with the input space of LLM. Specifically, $f_s$ and $f_p$ are fused with cross attention layer followed by residual connection:
\begin{equation}
f_s = f_s + \mathrm{MHCA}(\mathbf{K}=f_s, \mathbf{V}=f_s, \mathbf{Q}=f_p),
\end{equation}
    where $ \mathrm{MHCA}$ denotes Multi-Head Cross Attention. Finally, $f_s$ and $f_p$ are concatenated as the global feature $f$:
    \begin{equation}
    f = \mathrm{Concat}(f_s,f_p).
    \end{equation}
    For local images, we only use the semantic encoder to extract the feature for Interleaved Local Visual Coupling in sec.~\ref{method:ilvc}. During training, we freeze the semantic and pixel encoders, training only $\mathrm{MLP}_{p}$ and $\mathrm{MHCA}$.

\subsection{Interleaved Local Visual Coupling}
\label{method:ilvc}

    In LMMs, aligning local features with their corresponding local descriptions is essential for precise object comprehension.
    However, current models~\cite{lisa, omg-llava, glamm} typically extract the \texttt{<seg>} token, feed it into the decoder to generate masks, and use a binary \texttt{<mask, text>} pair for segmentation loss. This approach does not explicitly link local image regions to their corresponding text descriptions (see Fig.~\ref{fig:paradigm} (a)). However, human perception usually focuses on regions of interest before describing them.

    Inspired by this, we introduce Interleaved Local Visual Coupling (ILVC), a mask-region-text triplet interleaved training paradigm that assists in coupling local image regions with corresponding textual descriptions, as shown in Fig.~\ref{fig:paradigm} (b). First, we construct interleaved sequence data by leveraging the reference segmentation~\cite{refcoco,refcocog} and GranDf~\cite{glamm} datasets. Specifically, the global image is represented as $f_g$ and the N regions are represented as $\{f^{1}_{l}, f^{2}_{l}... f^{N}_{l}\}$ using the semantic encoder with a resolution of 448×448. Finally, we concatenate the encoded sequence into an interleaved feature sequence, which can be formulated as Eq.~\ref{sequence}.
    \begin{equation}
        S = \{f_g, T_{ins}; \textit{Seg}_1, f^{1}_{l}, T^{1}_{l}; ...;\textit{Seg}_n, f^{n}_{l}, T^{n}_{l}\},
    \label{sequence}
    \end{equation}
    where $T_{ins}$ denotes the text instruction, $\textit{Seg}_i$ denotes the $i_{th}$ \texttt{<seg>} token. As illustrated in Fig.~\ref{fig:arch}, we add special tokens\texttt{<p>} and \texttt{</p>} to explicitly identify the text descriptions corresponding to local image features.
    The \texttt{<seg>} token is decoded with a pixel decoder to predict the segmented map $M_i$.
    \begin{equation}
        M_i = \mathcal{D}(\mathcal{P}(I), \textit{Seg}_i),
    \end{equation}
    where $\mathcal{D}$ represents the pixel decoder. This supervises the model to learn instruction following and pixel segmentation capabilities.
    During the training process, we pre-extract the corresponding local regions from the original image using the ground truth (GT) mask.
    The local regions, after being encoded, are input into the LLM to predict the subsequent descriptions. The process is represented as:
    \begin{equation}
        T^{i}_{l} = \mathcal{\theta}(f_g, T_{ins}; \textit{Seg}_1, f^{1}_{l}, T^{1}_{l}; ...; \textit{Seg}_i, f^{i}_{l} ),
    \end{equation}
    where $\theta$ represents LLM. This approach allows the LLM to focus on the visual tokens derived from the sub-image, enabling it to generate more accurate and detailed descriptions of the specified area and thereby minimizing the occurrence of hallucinations. In this training paradigm, we successfully couple the local sub-images and the corresponding text descriptions, establishing the connection between masks, local images, and corresponding descriptions. During the inference process, we extract the corresponding image based on the decoded mask instead of the ground truth.

  \begin{table*}[]
    \centering  
    \scalebox{0.65}{
    \begin{tabular}{c|cccccccc|cccccccc}
    \toprule
                       &                          &                       &                         &                         &                         &                        &                             &                             & \multicolumn{3}{c}{RefCOCO} & \multicolumn{3}{c}{RefCOCO+} & \multicolumn{2}{c}{RefCOCOg} \\
    \multirow{-2}{*}{} & \multirow{-2}{*}{VizWiz} & \multirow{-2}{*}{GQA} & \multirow{-2}{*}{VQAv2} & \multirow{-2}{*}{OKVQA} & \multirow{-2}{*}{SciQA} & \multirow{-2}{*}{POPE} & \multirow{-2}{*}{MMB-en} & \multirow{-2}{*}{MMB-cn} & Val     & TestA   & TestB   & Val     & TestA    & TestB   & Val           & Test         \\ \midrule
    \rowcolor[HTML]{F2F3F5} \multicolumn{17}{l}{\textit{Segmentation-supporting}} \\
    LISA-7B~\cite{lisa}            & -                        & -                     & -                       & -                       & -                       & 0.0                      & 0.4                         & -                           & 74.1    & 76.5    & 71.1    & 62.4    & 67.4     & 56.5    & 66.4          & 68.5         \\
    LISA-GLEE~\cite{glee}         & -                        & -                     & -                       & -                       & -                       & -                      & -                           & -                           & 76.4    & 78.2    & 73.8    & 67.3    & 71.3     & 62.3    & 71.6          & 72.4         \\
    PixelLM-13B~\cite{pixellm}        & -                        & -                     & -                       & -                       & -                       & 0.0                      & 17.4                        & -                           & 73.0      & 76.5    & 68.2    & 66.3    & 71.7     & 58.3    & 69.3          & 70.5         \\
    GSVA(ft)~\cite{gsva}        & -                        & -                     & -                       & -                       & -                       & -                      & -                           & -                           & 77.2    & 78.9    & 73.5    & 65.9    & 69.6     & 59.8    & 72.7          & 73.3         \\
    GLaMM†~\cite{glamm}         & -                        & -                     & -                       & -                       & -                       & -                      & -                           & -                           & 79.5    & 83.2    & 76.9    & 72.6    & \underline{78.7}     & 64.6    & 74.2          & 74.9         \\
    PSALM~\cite{psalm}              & -                        & -                     & 62.3                    & -                       & 64.9                    & 80.3                   & 52.5                        & -                           & \textbf{83.6}    & \textbf{84.7}    & \textbf{81.6}    & \underline{72.9}    & 75.5     & \underline{70.1}    & 73.8          & 74.4         \\
    OMG-LLaVA-8B~\cite{omg-llava}       & -                        & -                     & -                       & -                       & 57.8                    & 80.0                     & 47.9                        & -                           & 75.6    & 77.7    & 71.2    & 65.6    & 69.7     & 58.9    & 70.7          & 70.2         \\
    OMG-LLaVA-8B(ft)         & -                        & -                     & -                       & -                       & -                       & -                      & -                           & -                           & 78.0    & 80.3    & 74.1    & 69.1    & 73.1     & 63.0    & 72.9          & 72.9         \\
    \midrule
    \rowcolor[HTML]{F2F3F5} \multicolumn{17}{l}{\textit{Comprehension-only}} \\
    LLaVA-Phi-2.7B~\cite{llava-phi}     & 35.9                     & -                     & 71.4                    & -                       & 68.4                    & 85.0                     & 59.8                        & -                           & -       & -       & -       & -       & -        & -       & -             & -            \\
    Mini-Gemini-2B~\cite{mini-gemini}     & -                        & -                     & -                       & -                       & -                       & -                      & 59.8                        & -                           & -       & -       & -       & -       & -        & -       & -             & -            \\
    MiniCPMV2-2.4B~\cite{minicpm}     & -                        & -                     & -                       & -                       & 80.7                    & 86.3                   & 69.1                        & {\color[HTML]{1F2937} 66.5} & -       & -       & -       & -       & -        & -       & -             & -            \\
    Qwen-VL-10B~\cite{qwen-vl}       & 35.2                    & 59.3                   & 79.5                    & 58.6                   & 67.1                    & 70.0                  & 32.2                        & 7.8                        & -       & -       & -       & -       & -        & -       & -             & -  \\
    LLaVA-Next-Llama-8B~\cite{llava-next}       & -                    & -                   & -                    & -                   & 73.7                    & 87.1                  & 74.8                        & 70.1                        & -       & -       & -       & -       & -        & -       & -             & -  \\
    Cambrian-1-8B~\cite{cambrian}       & -                    & \underline{64.6}                   & -                    & -                   & 80.4                    & 86.4                  & 68.2                        & -                        & -       & -       & -       & -       & -        & -       & -             & -  \\
    Monkey-10B~\cite{monkey}       & 61.2                    & 60.7                   & 80.3                    & \underline{61.3}                   & 69.4                    & 83.7                  & 59.6                        & 54.7                        & -       & -       & -       & -       & -        & -       & -             & -  \\
    mPLUG-Owl3-8B~\cite{mplug-owl3}       & 63.5                    & \textbf{65.0}                   & \textbf{82.1}                    & 60.1                   & -                    & 88.2                  & 77.6                        & 74.3                        & -       & -       & -       & -       & -        & -       & -             & -  \\

    InternVL2-2B~\cite{internvl2}       & 47.4                     & 61.0                    & 76.6                    & 53.2                    & 94.1                    & \underline{88.3}                   & 73.2                        & 70.9                        & -       & -       & -       & -       & -        & -       & -             & -  \\
    InternVL2-8B~\cite{internvl2}       & 62.9                     & 63.2                   & 79.8                   & \textbf{62.9}                  & \underline{97.1}                    & 86.9                 & \textbf{81.7}                        & \textbf{81.2} & -       & -       & -       & -       & -        & -       & -             & -  
    \\ \midrule
    \rowcolor[HTML]{F2F3F5} \multicolumn{17}{l}{
    \textit{Ours}} \\
    LIRA-2B            & \underline{67.8}                     & 61.1                  & 77.2                    & 53.7                    & 95.0                      & \textbf{89.1}                   & 74.0                          & 71.7                        & 79.5    & 81.9    & 76.0      & 72.6    & 77.4     & 66.1    & \underline{75.4}          & \underline{75.1} 
    \\
    LIRA-8B            & \textbf{71.5}                     & 63.5                  & \underline{80.4}                    & \textbf{62.9}                    & \textbf{97.3}                    & 88.1                   & \underline{81.1}                        & \underline{80.5}                        & \underline{81.8}    & \underline{83.4}    & \underline{78.1}    & \textbf{76.3}    & \textbf{81.1}     & \textbf{70.5}    & \textbf{78.4}          & \textbf{78.2}         \\ \bottomrule
    
    \end{tabular}}
    \caption{The comprehensive comparison of LIRA and other LMMs regarding reasoning capability and pixel-level performance.
    We test the datasets of comprehension: VizWiz~\cite{vizwiz}, GQA~\cite{gqa}, VQAv2~\cite{vqav2}, OKVQA~\cite{okvqa}, SciQA~\cite{scienceqa}, POPE~\cite{pope}, MMB~\cite{mmbench}. And as for segmentation tasks, we choose RefCOCO~\cite{refcoco}, RefCOCO+~\cite{refcoco}, RefCOCOg~\cite{refcocog}. Underline represents the second best, ``ft" indicates finetuning on the referring expression datasets, † indicates that the method used the GranD dataset~\cite{glamm} for pretraining, which is significantly larger than the datasets used by other methods.}
    
    \label{tab:main}
    \end{table*}

    \subsection{Training Objectives}
    The training process of LIRA can be divided into two stages. In the first stage, only the $\mathrm{MLP}_p$ and the text projector are trained, with the pixel encoder’s features being used solely for vision-language alignment. In the second stage, in addition to fine-tuning the $\mathrm{MLP}_p$ and the text projector, we also fine-tune the $\mathrm{MHCA}$ for feature fusion. Moreover, we employ LoRA to fine-tune the LLM. LIRA is trained end-to-end by combining text generation loss and segmentation loss. The loss function of the second stage is expressed as follows:
    
    \begin{equation}
    L = L_{\text{text}} + \alpha L_{\text{mask}},
    \end{equation}
    where $L_{mask}$ indicates the mask loss, which comprises a pixel-level Cross-Entropy (CE) loss and Dice loss, $L_{\text{text}}$  indicates the next token prediction loss of LLM. The $\alpha$ is a hyperparameter balancing the weight of the mask loss.

\section{Experiments}

%

\begin{table*}
    \centering
    \small
    \begin{tabular}{c|cccc|cccc}
    \toprule
    \multirow{2}{*}{Methods} & \multicolumn{4}{c|}{Val}     & \multicolumn{4}{c}{Test}     \\
                             & METEOR & CIDEr & ap50 & mIoU & METEOR & CIDEr & ap50 & mIoU \\ \midrule
    Kosmos-2~\cite{kosmos2}                 & \textbf{16.1}   & 27.6  & 17.1 & 55.6 & \textbf{15.8}   & 27.2  & 17.2 & 56.8 \\
    LISA-7B~\cite{lisa}                     & 13.0     & 33.9  & 25.2 & 62.0   & 12.9   & 32.2  & 24.8 & 61.7 \\
    OMG-LLaVA-8B~\cite{omg-llava}                 & 13.8   & 36.2  & 26.9 & 64.6 & 13.5   & 33.1  & \textbf{26.1} & 62.8 \\
    LIRA-8B                     & 14.1     & \textbf{38.4}  & \textbf{28.1} & \textbf{64.9} & 13.7   & \textbf{36.4}  & 25.6 & \textbf{63.1} \\ \bottomrule
    \end{tabular}
    \caption{Results on Grounded Conversation Generation.}
    
    \label{tab:gcg}
    \end{table*}

\subsection{Implementation Details}
    We use InternVL2-2B, InternVL2-8B and OMG-Seg~\cite{omgseg} as our baseline to develop LIRA. Our model's semantic encoder and large language model (LLM) are built upon InternVL2, while the pixel encoder and pixel decoder are adapted from OMG-Seg. 
    In the first stage, we align the pixel encoder with the LLM, with an initial learning rate set to 1e-3. During the instruction-tuning stage, we train the LLM using LoRA~\cite{lora} with a rank of 128 for the 2B model and a rank of 256 for the 8B model, setting the learning rate to 2e-5. The global batch size is set to 128. The entire instruction-tuning process for the 2B model takes approximately 20 hours using 8 A800 GPUs.
\subsection{Datasets Setup}
    Referring to the same data construction pipeline of OMG-LLaVA, we have devised a two-stage data. In the first stage, we utilize 557k detailed caption datasets from~\cite{monkey, sharegpt4v} for pretraining. In the instruction tuning stage, we use 411k data from comprehension dataset~\cite{llava,scienceqa,vqav2,dvqa,aokvqa,vizwiz,ai2d,okvqa,chartqa,gqa},  and 374k from segmentation datasets ~\cite{refcoco,refcocog,glamm} for training. The details can be found in the Appendix. 

    \begin{table}
    \resizebox{\linewidth}{!}{
    \begin{tabular}{c|c|cccccc}
    \toprule
    \multirow{3}{*}{Model} & \multirow{3}{*}{zero-shot} & \multicolumn{6}{c}{Generalized Referring Segmentation}                          \\
                           &                            & \multicolumn{2}{c}{val} & \multicolumn{2}{c}{testA} & \multicolumn{2}{c}{testB} \\
                           &                            & cIoU       & gIoU       & cIoU         & gIoU       & cIoU        & gIoU        \\ \midrule
    ReLA~\cite{rela}                   & x                          & 62.4       & 63.6       & 69.3         & 70.0         & 59.9        & 61.0          \\
    LISA~\cite{lisa}                   & x                          & 38.7       & 32.2       & 52.6         & 48.5       & 44.8        & 39.7        \\
    LISA(ft)~\cite{lisa}                   & x                          & 61.7       & 61.6       & 69.2         & 70.1       & 60.3        & 61.3        \\
    GSVA~\cite{gsva}                   & x                          & 61.7       & 63.3       & 69.2         & 70.1       & 60.3        & 61.3        \\
    GSVA(ft)~\cite{gsva}                   & x                          & 63.3       & 66.5       & 69.9         & 71.1       & 60.5        & 62.2        \\ \midrule
    LaSanA~\cite{lasagna}                 & \checkmark                          & 38.1       & 32.4       & 50.4         & 47.3       & 42.1        & 38.9        \\
    OMG-LLaVA~\cite{omg-llava}               & \checkmark                         & 39.3          & 36.1          & 52.4            & 50.1          & 43.7          &  42.2           \\
    LIRA-8B                   & \checkmark                         & 40.9       & 36.7       & 52.4         & 50.4       & 44.9        & 42.4        \\ \bottomrule
    \end{tabular}}
    \caption{Comparison with other methods on gRefCOCO~\cite{grefcoco}.}
    \label{tab:grefcoco}
    \end{table}

\subsection{Comparison with other LMMs}
We evaluate our model by testing it across a wide range of
standard vision-language tasks. It is worth noting that, unlike previous methods that may achieve improvements through task-specific fine-tuning after the instruction tuning stage, we only perform one instruction tuning stage. 
\textbf{Comprehension and Referring Expression Segmentation Benchmarks.}
We evaluated our model, LIRA, on several commonly used benchmarks for Comprehension and Referring Expression Segmentation. The results, as shown in Tab.~\ref{tab:main}, demonstrate that LIRA achieves improved performance in both comprehension and segmentation tasks. Specifically, we obtained an average accuracy of 78.2\% across eight comprehension benchmarks, showcasing strong overall performance. On the RefCOCOg dataset, our 8B model achieved accuracies of 78.4\% on the validation set and 78.2\% on the test set. Furthermore, LIRA notably enhances the segmentation capabilities, achieving top performance in 5 out of 8 standard evaluation settings across the RefCOCO, RefCOCO+, and RefCOCOg datasets. It is worth noting that we suspect that PSALM uses 100 queries to predict the mask of the same object and selects the highest-scoring one, which may gain advantages on certain datasets.

        \begin{table*}[]
    \centering
    \scalebox{0.9}{
    \begin{tabular}{cc|cccc|cccccc}
    \toprule  
    \multirow{2}{*}{LLM} & \multirow{2}{*}{SEFE} & \multirow{2}{*}{VizWiz} & \multirow{2}{*}{VQAv2} & \multirow{2}{*}{MMB-en} & \multirow{2}{*}{MMB-cn}  & \multicolumn{3}{c}{RefCOCO} & \multicolumn{3}{c}{RefCOCO+}\\
                      &                         &                        &                       &                         &      & Val     & TestA   & TestB   & Val     & TestA    & TestB     \\ \midrule
    InternLM2-1.8B &  -    & 62.7                    & 71.1                   & 66.7                  & 63.9                                   & 77.2    & 78.6    & 72.9    & 68.6      & 72.2     & 60.9            \\
    InternLM2-1.8B &   \checkmark   & 67.0                    & 76.1                   & 73.8                  & 70.3                                     & 79.4    & 81.2    &  76.1  & 72.7    & 77.5     & 66.1      \\
    
    InternLM2.5-7B &  -            &     65.7                & 75.1                   & 76.2                  & 74.6                                      & 79.3    & 81.2    & 75.3      & 72.3    & 76.7    &  65.7    \\
     InternLM2.5-7B &  \checkmark  & 71.1                    & 80.0                   & 80.8                  & 80.1                                      & 81.7    & 83.3    & 77.8      & 76.4    & 80.8     & 70.7 
    \\ \bottomrule
    \end{tabular}
    }
    \caption{Ablation study on the effect of SEFE on comprehension and segmentation tasks. }
    \label{tab:sefe_effect}
    \end{table*}

\textbf{Grounded Conversation Generation.}
As shown in the Tab.~\ref{tab:gcg}, we have achieved excellent results without specific fine-tuning. Although our performance on the METEOR metric is lower than that of Kosmos-2~\cite{kosmos2}, which uses nearly ten times more data, we have surpassed existing models in other metrics, demonstrating the strong comprehension and segmentation capabilities of our model. 
Especially on the CIDER metric, we achieve scores of 38.4\% and 36.4\% on the validation and test sets, exceeding the previous method by 2.75\%.
We also qualitatively demonstrate the effectiveness of our method in Sec.~\ref{visual} and the Appendix.

\textbf{Generalized Referring Expression Segmentation.}
We evaluate LIRA on the gRefCOCO~\cite{grefcoco} benchmark, which includes multiple segmentation targets and even cases without targets. In practice, to maintain consistency with previous results, we utilize the testing format of RefCOCO. As shown in Tab.~\ref{tab:grefcoco}, LIRA, trained on  374k segmentation instances, shows promising zero-shot performance, even surpassing LISA trained on gRefCOCO.

\subsection{Ablation Study}
\label{exp:ab}
 To fully validate the effectiveness of our design, we conduct ablation studies on the key designs of LIRA.

\textbf{Effectiveness of SEFE.} To validate the effectiveness of SEFE, we conducted ablation experiments using the InternLM2-1.8B ~\cite{internlm2} and InternLM2.5-7B~\cite{internlm2} backbones. As shown in Tab.~\ref{tab:sefe_effect}, incorporating SEFE led to an average improvement of 5.7\% on understanding tasks and 3.8\% on segmentation tasks when using the InternLM2-1.8B backbone. Similarly, with the InternLM2.5-7B backbone, we observed an average improvement of 5.1\% on understanding tasks and 3.4\% on segmentation tasks.
This improvement can likely be attributed to SEFE's ability to enhance the model's visual comprehension capabilities, enabling it to interpret visual instructions more effectively and thereby improve segmentation accuracy. 

\begin{figure}[]
  \centering
   \includegraphics[width=1.0\linewidth]{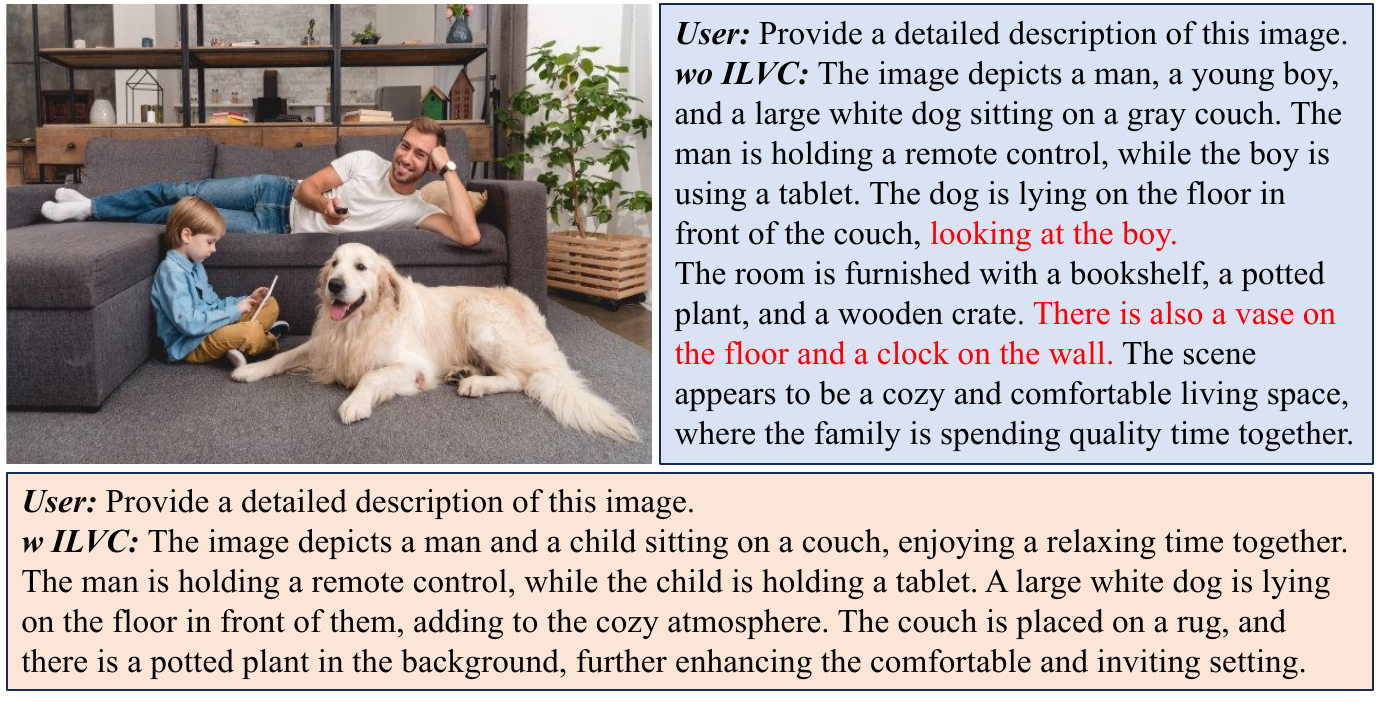}
   \caption{Comparing the hallucination of \textit{wo/w} ILVC, with hallucination content highlighted in red.}
   \label{fig:hul}
\end{figure}



\textbf{Effectiveness of ILVC.} Building on the implementation of SEFE, we conducted ablation studies to explore the effectiveness of further integrating ILVC. As shown in Tab.~\ref{tab:fgia_hu}, we achieved consistent improvements in reducing hallucinations by adopting ILVC, which aligns local visual features with their corresponding local descriptions. Specifically, the hallucinations of ChairS~\cite{chair} decreased by 3.0\% and 4.8\% for the 1.8B and 7B LLMs, respectively. To further illustrate the impact of ILVC on hallucinations, we provide detailed captions generated by models with and without ILVC in Fig.~\ref{fig:hul}. By integrating ILVC, the generated caption is more accurate, and many non-existent objects and attributes have been eliminated from the images, such as ``the dog is looking at the boy" and ``there is also a vase on the floor and a clock on the wall".

\textbf{Effectiveness of co-training with both comprehension and segmentation data.} As indicated in Tab.~\ref{tab:egmf}, co-training LIRA with both comprehension and segmentation data leads to only a slight decrease (-0.2\%) compared to training solely on comprehension data, surpassing the previous leading method, OMG-LLaVA (-14.3\%), across five comprehension datasets. 
It is worth noting that on the MME, MMBench, and POPE datasets, our method maintained or even improved upon the original performance, demonstrating a significant advantage compared to a drop of nearly 15\% with OMG-LLaVA.
This demonstrates that our model retains its original comprehension capabilities while possessing segmentation abilities.

\textbf{Effectiveness of LIRA on enhancing comprehension abilities.}
As shown in Tab.~\ref{tab:egmf} and Tab.~\ref{tab:main}, we enhance segmentation capabilities while slightly improving comprehension accuracy (75.2\% vs. 74.8\%) compared to IntenVL2 through LIRA. 
Notably, in terms of hallucination reduction, our model shows clear performance improvements across the Chair, POPE, and TinyLVLM datasets, achieving consistent gains with both the 2B and 8B models.



\begin{table}
\centering
\scalebox{0.62}{
\begin{tabular}{c|ccccc}
\toprule
Methods & MME & MMBench & SEED-Bench & POPE & AI2D \\
\midrule
\multicolumn{6}{c}{Training only with comprehension dataset} \\
\midrule
LLaVA 1.5 & 1422/267 & 68.5 & 65.9 & 86.7 & 56.6 \\
OMG-LLaVA & 1448/282 & 67.5 & 68.9 & 89.7 & 61.7 \\
LIRA-2B  & 1455/409 & 74.0 & 71.1 & 89.1 & 75.9 \\
\midrule
\multicolumn{6}{c}{Co-training with comprehension dataset and segmentation datasets} \\
\midrule
LISA & 1/1 & 0.4 & - & 0.0 & 0.0 \\
PixelLM & 309/135 & 17.4 & - & 0.0 & 0.0 \\
LaSagnA & 0/0 & 0.0 & - & 0.0 & 0.0 \\
GLaMM & 14/9 & 36.8 & - & 0.94 & 28.2 \\
OMG-LLaVA & 1177/235 (-318) & 47.9 (-19.6) & 56.5 (-12.4) & 80.0 (-9.7) & 42.9 (-18.8) \\
LIRA-2B  & 1429/440 (+5) & 74.0 (+0) & 71.0 (-0.1) & 89.1 (+0) & 75.0 (-0.9) \\
\bottomrule
\end{tabular}}
\caption{Performance on image-level benchmarks \textit{wo/w} segmentation datasets. Partial results are excerpted from OMG-LLaVA~\cite{omg-llava}.}
\label{tab:egmf}
\end{table}

\subsection{Visualization}
\label{visual}
\begin{figure*}[]
  \centering
   \includegraphics[width=0.9\linewidth]{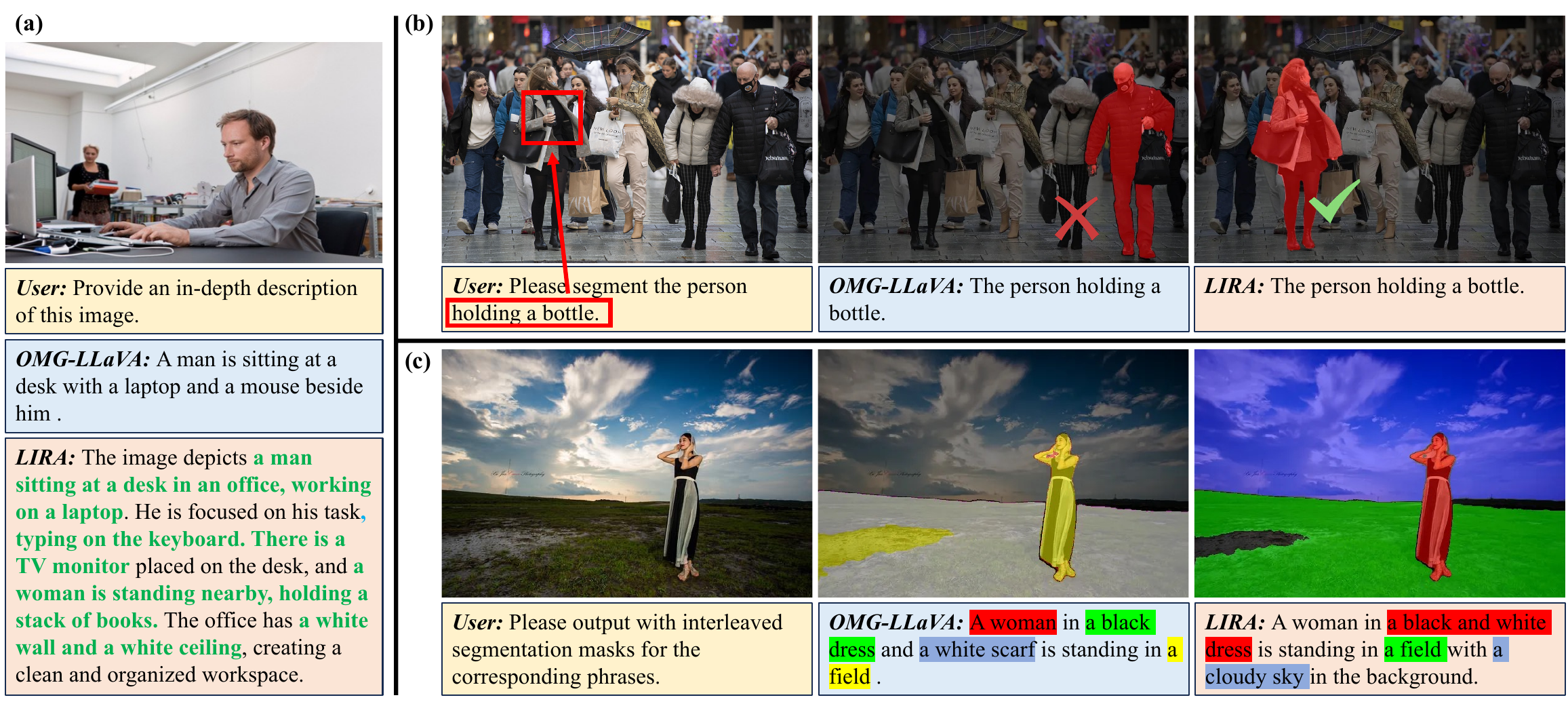}
   \caption{Comparison with OMG-LLaVA in image caption, referring expression segmentation, and grounded conversation generation. The green text in (a) corresponds to objects present in the image. More examples can be found in the appendix.}
   \label{fig:visual}
\end{figure*}

We evaluate LIRA in image captioning, referring segmentation, and grounded conversation segmentation tasks compared with OMG-LLaVA. As demonstrated in Fig.~\ref{fig:visual} (a), our model generates richer detailed descriptions. Furthermore, our model accurately captures the features of the object to be segmented, ``holding a bottle'', thereby achieving correct content segmentation, as shown in Fig.~\ref{fig:visual} (b). Additionally, in the GCG task, our model establishes finer-grained alignment relationships, allowing for a better generation of masks corresponding to the descriptions in Fig.~\ref{fig:visual} (c).


\section{Discussion}
\label{discussion}

\begin{table}
\scalebox{0.82}{
\begin{tabular}{cc|cccc}
\toprule
\multirow{2}{*}{LLM} & \multirow{2}{*}{ILVC} & \multicolumn{4}{c}{Hallucination} \\
 &  & ChairS$\downarrow$ & ChairI$\downarrow$ & POPE & TinyLVLM \\ \midrule
 
 InternLM2-1.8B & - & 32.2 & 11.1 & 88.8 & 90.7 \\

 InternLM2-1.8B & \checkmark & \textbf{29.2 } & \textbf{10.4}& \textbf{89.1} & \textbf{92.0}  \\ \midrule

InternLM2.5-7B & - & 43.6 & 12.1 & 87.7 & 88.7  \\

InternLM2.5-7B &  \checkmark & \textbf{38.8} & \textbf{10.6}& \textbf{88.1 } & \textbf{89.7}  \\

\bottomrule
\end{tabular}}
\caption{Ablation study on the influence of ILVC in mitigating large multi-modal model hallucinations. For TinyLVLM~\cite{tinylvlm}, we select the Object Hallucination component to evaluate hallucinations. Bold font represents the best performance, $\downarrow$ indicates that a smaller number is better.}
\label{tab:fgia_hu}
\end{table}

\begin{figure}[]
  \centering
   \includegraphics[width=0.9\linewidth]{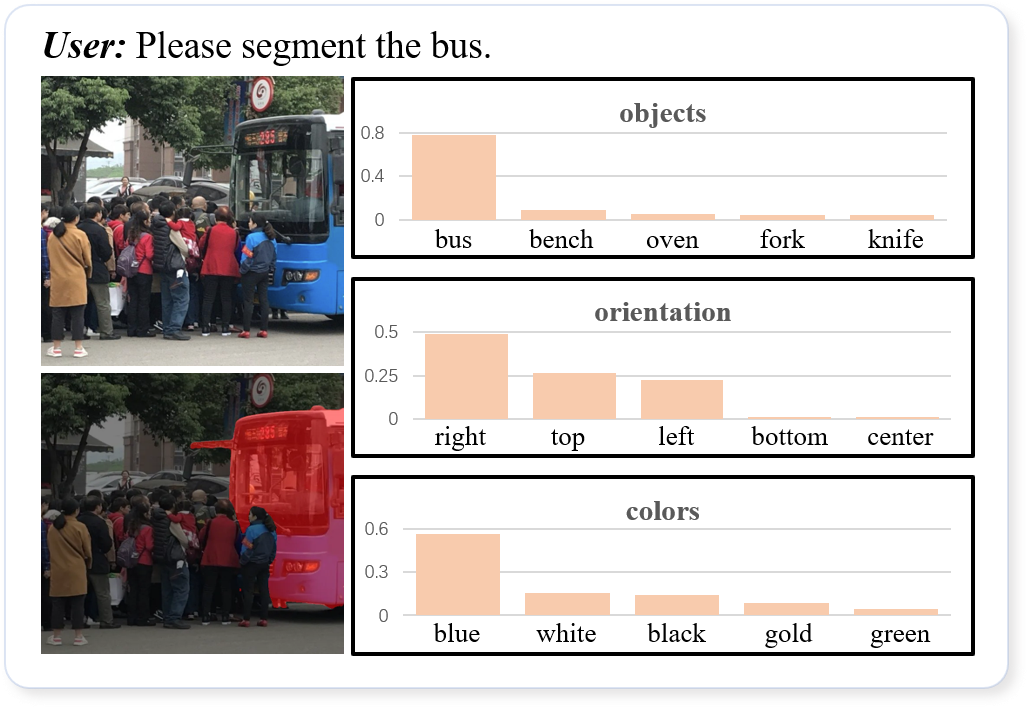}
   \caption{Visualization of the logits values for the \texttt{<seg>} token. The bar chart on the right highlights the top five tokens with the highest logits in terms of objects, orientations, and colors.}
   \label{fig:blue_bus}
\end{figure}

    In an illustrative experiment (Fig.~\ref{fig:analysis}), we analyze the \texttt{<seg>} token used for decoding objects, such as ``the red bus closest to the white car.” Logits analysis reveals that a higher probability of predicting ``right" corresponds to right-side segmentation, while ``left" correlates with left-side segmentation.In Fig.~\ref{fig:blue_bus}, we further extract the top five tokens from the \texttt{<seg>} token based on the highest logits concerning objects, orientations, and colors. We find that the highest logits within the \texttt{<seg>} token effectively represent the correct attributes of the objects.

    To further investigate quantitatively, we extracted information on the color and orientation attributes of objects by analyzing different descriptions of the same object in RefCOCO. Based on the extracted attribute information, we introduce the Attributes Evaluation (AttrEval) dataset, comprising segmentation and question-answering tasks. The segmentation task still follows the original reference expression segmentation task, while the QA task involves asking questions based on the attributes of an object to determine if it possesses that attribute. For segmentation, we use RefCOCO data requiring attribute inferring. The QA task assesses spatial attributes (e.g., ``Is xxx on the right/left?'') and evaluates whether predictions align with top logit scores ($Acc_1$ and $Acc_3$) and VQA accuracy. Results in Tab. \ref{tab:discuss} demonstrate ILVC's improved semantic alignment, achieving scores of 25.7\%, 38.8\%, and 58.6\% in $Acc_1$, $Acc_3$, and VQA accuracy.

    \textbf{Limitation.} Although introducing ILVC does not significantly increase the overhead during training, there are still some costs associated with inference. For QA or caption tasks, if mask decoding is not required, the inference efficiency of these tasks will not be affected. For more complex tasks such as Grounded Conversational Generation, there is an additional inference overhead of approximately 15\%. Our proposed method achieves a top-1 accuracy of 25.7\% for logits in AttrEval and 58.6\% in question answering, indicating considerable room for improvement in attribute inference. These results highlight the need for further exploration to enhance semantic understanding and alignment capabilities in the future.


      \begin{table}[]
    \centering
    \begin{tabular}{c|ccc}
    \toprule
     & $Acc_{1}$ & $Acc_{3}$ & VQA acc \\ \midrule
    OMG-LLaVA & 6.3 & 22.3 & 41,1 \\
    LIRA-8B & \textbf{25.7} & \textbf{38.8} & \textbf{58.6} \\
    LIRA-8B \textit{wo} ILVC & 22.3 & 36.4 &  55.1  \\ \bottomrule
    \end{tabular}
    \caption{The results on the AttrEval.}
    \label{tab:discuss}
    \vspace{-1em}
    \end{table}

\section{Conclusion}

In this paper, we propose LIRA, a new paradigm that can enable the segmentation ability of a comprehension-based LMM while reducing hallucinations. The experiments demonstrate improvements in our method across multiple benchmarks, including comprehension, segmentation, grounded conversation generation, and hallucination. Furthermore, we introduce the AttrEval dataset, which evaluates the model’s comprehension of object attributes. Through the analysis, we find that a deeper understanding of image attributes can enhance model performance. In the future, further exploration of textual-visual correlations and attribute inference is warranted.

\section*{Acknowledgements}

This work was supported by the NSFC (U2341227 and 62206104).

{
    \small
    \bibliographystyle{ieeenat_fullname}
    \bibliography{main}
}

\newpage
\appendix
\onecolumn
\section{Summary of the Instruction Tuning Data}
We provide the detailed composition of our instruction tuning data in Tab.~\ref{tab:data}, which contains a total of 785K samples.
For the comprehension task, we select eleven widely used datasets, comprising a total of 411K samples. These include TextVQA~\cite{textvqa}, which requires the model to answer questions by reading and reasoning about text within images; DVQA~\cite{dvqa}, which focuses on processing words and answers related to bar charts; and ChartQA~\cite{chartqa}, which involves visual and logical reasoning about charts. Additionally, LLaVA150K~\cite{llava} is a GPT-generated dataset for multimodal instruction-following tasks, while ScienceQA~\cite{scienceqa} and AI2D~\cite{ai2d} are centered around science topics. VQAV2~\cite{vqav2} targets open-ended visual question answering on natural images, and OKVQA~\cite{okvqa} extends this by requiring additional world knowledge. AOKVQA~\cite{aokvqa} further incorporates commonsense reasoning to answer questions about scenes. VizWiz~\cite{vizwiz} is designed to answer questions posed by blind individuals in real-world scenarios. Following LLaVA , we incorporated the prompt, ``When the provided information is insufficient, respond with `Unanswerable,'" during both training and inference.  Finally, GQA~\cite{gqa} is a dataset focused on real-world visual reasoning and compositional question answering.
For the segmentation task, we select data from two key tasks: referring expression segmentation (RefSeg) and grounded conversation generation (GCG). In the RefSeg task, which involves object segmentation based on natural language descriptions, we use the RefCOCO~\cite{refcoco}, RefCOCO+~\cite{refcoco}, and RefCOCOg~\cite{refcocog} datasets, totaling 168k samples. For the GCG task, which aims to generate detailed image descriptions with corresponding masks for the phrases, we use the GranDf~\cite{glamm} dataset, containing 206K samples.

\begin{table}[h]
\centering
\scalebox{0.9}{
\begin{tabular}{l|l|l|l}
\toprule
Task                                     & Dataset   & Description                                                                                                                & Samples \\ \midrule
                                         & TextVQA~\cite{textvqa}   & VQA involving reading and reasoning about text                                                                             & 15k     \\
                                         & LLaVA150k~\cite{llava} & GPT-generated multimodal instruction-following data                                      & 157k    \\
                                         & ScienceQA~\cite{scienceqa} & Multimodal multiple choice VQA on science topics                                                                           & 15k     \\
                                         & VQAV2~\cite{vqav2}     & Open-ended VQA about natural images                                                                                        & 60k     \\
                                         & DVQA~\cite{dvqa}     & Understanding Data Visualizations via Question Answering                                                                   & 10k     \\
                                         & AOKVQA~\cite{aokvqa}    & A Benchmark for Visual Question Answering using World Knowledge                                                            & 30k     \\
                                         & VizWiz~\cite{vizwiz}    & Answering visual questions from blind people                                                                               & 10k     \\
                                         & AI2D~\cite{ai2d}      & Multiple choice VQA on science diagrams                                                                                    & 30k     \\
                                         & OKVQA~\cite{okvqa}     & VQA involving world knowledge on natural images                                                                            & 9k      \\
                                         & CharQA~\cite{chartqa}    & VQA on charts with visual and logical reasoning                                                                            & 15k     \\
\multirow{-11}{*}{Comprehension}         & GQA ~\cite{gqa}       & Real-world visual reasoning and compositional question answering                                                           & 60k     \\ \midrule
                                         & RefCOCO~\cite{refcoco}   &                                                                                                                            & 51k     \\
                                         & RefCOCO+~\cite{refcoco}  &                                                                                                                            & 51k     \\
\multirow{-3}{*}{RefSeg} & RefCOCOg~\cite{refcocog}  & \multirow{-3}{*}{Object segmentation based on natural language descriptions}                 & 66k     \\ \midrule
GCG         & GranDf~\cite{glamm}    & 
Generate a detailed image description with corresponding masks for the phrases & 206k    \\ \midrule
Total                                    & -         &      -                                                                                                                      & 785k    \\ \bottomrule
\end{tabular}}
\caption{Details of the Instruction Tuning Data.}
\label{tab:data}
\end{table}

\newpage
\section{More Visualization Results}

\begin{figure*}[h]
      \centering
       \includegraphics[width=0.95\linewidth]{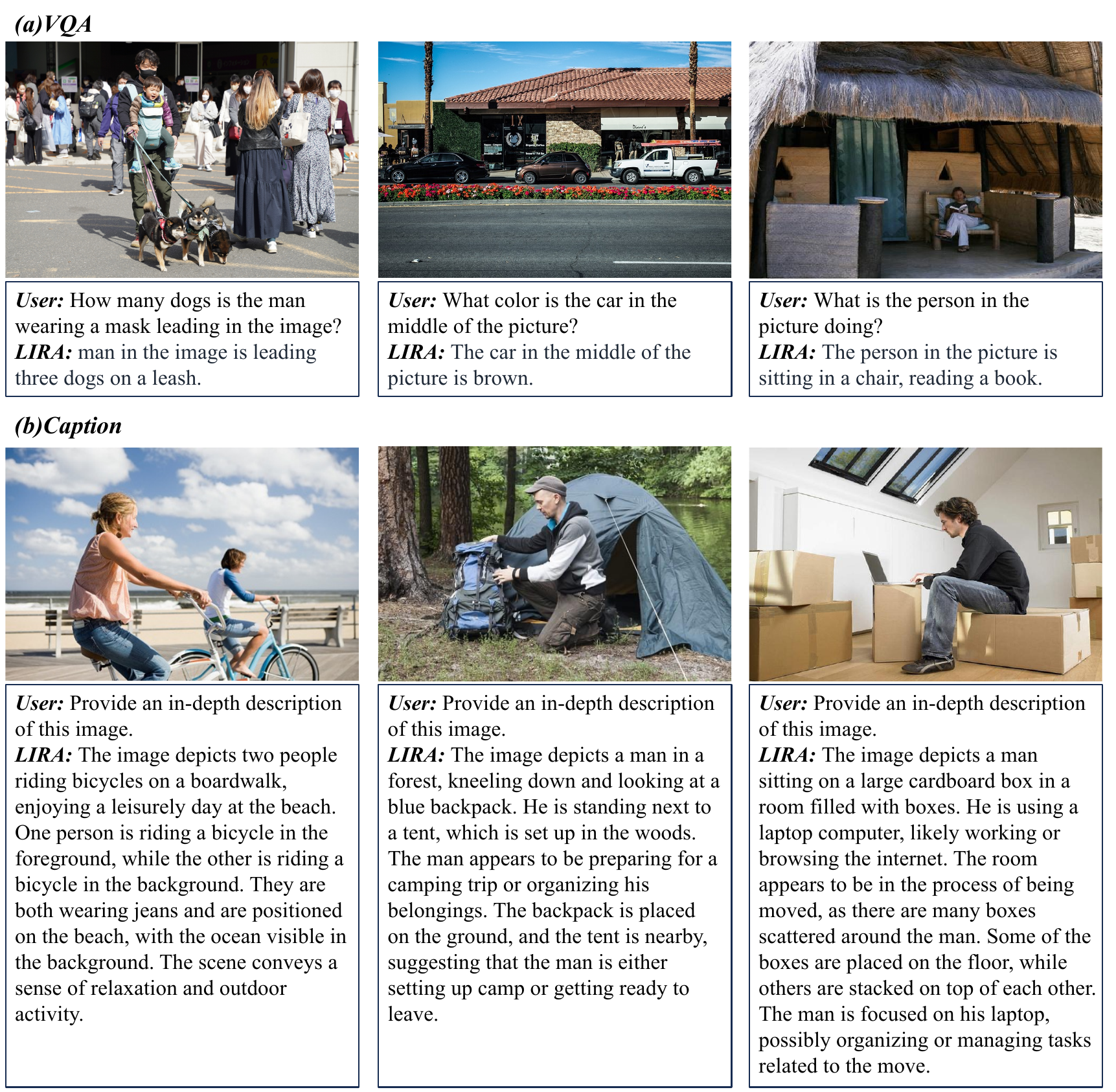}
       \caption{The visualization results of the VQA and Image Caption tasks.}
       \label{fig:vqa}
    \end{figure*}

\begin{figure*}[t]
      \centering
       \includegraphics[width=0.95\linewidth]{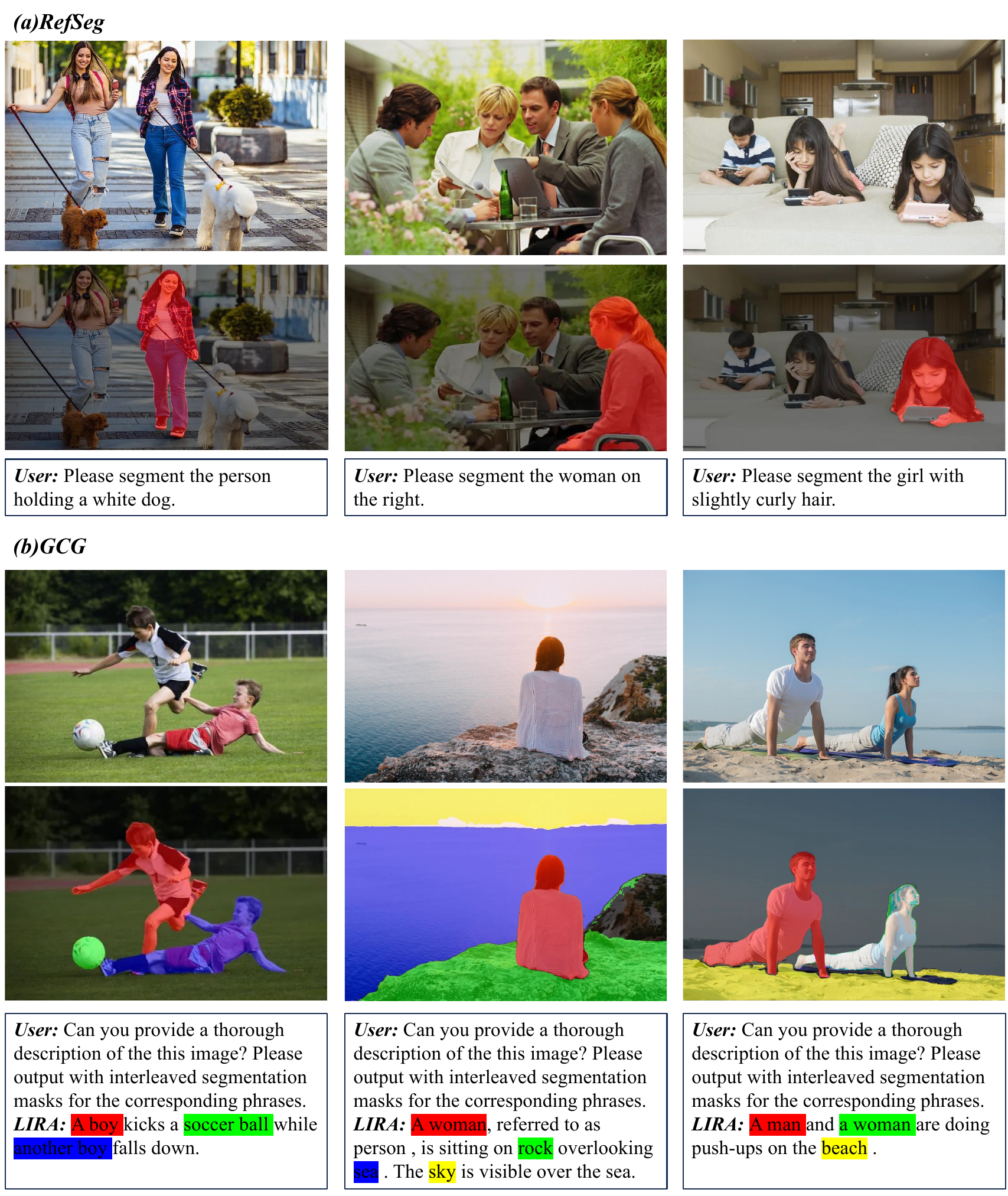}
       \caption{The visualization results of the referring expression segmentation (RefSeg) and grounded conversation generation (GCG) tasks.}
       \label{fig:refseg}
    \end{figure*}

We present additional visualization results, where Fig.~\ref{fig:vqa} and Fig.~\ref{fig:refseg} showcase LIRA’s capabilities across various tasks. 
As shown in Fig.~\ref{fig:vqa}, LIRA demonstrates exceptional scene understanding capabilities, accurately analyzing image content, responding to complex queries, and providing clear and detailed scene descriptions. For instance, LIRA correctly identifies that the man in the first image is walking three dogs and that the car in the center of the second image is brown, showcasing precise recognition of object attributes. Additionally, LIRA not only provides an accurate summary of the scene but also captures the underlying emotions, such as the relaxed atmosphere of the seaside cycling depicted in the first image.
As shown in Fig.~\ref{fig:refseg}, LIRA is capable of understanding the attributes of objects specified in the instructions, such as ``person holding a white dog," ``woman on the right," and ``girl with slightly curly hair," and accurately segmenting the targets. In the GCG task, LIRA is capable of generating descriptions of the image and accurately segmenting the objects mentioned in the descriptions.

\newpage

\section{Comparing Hallucination \textit{wo/w} ILVC}

\begin{figure*}[t]
      \centering
       \includegraphics[width=0.95\linewidth]{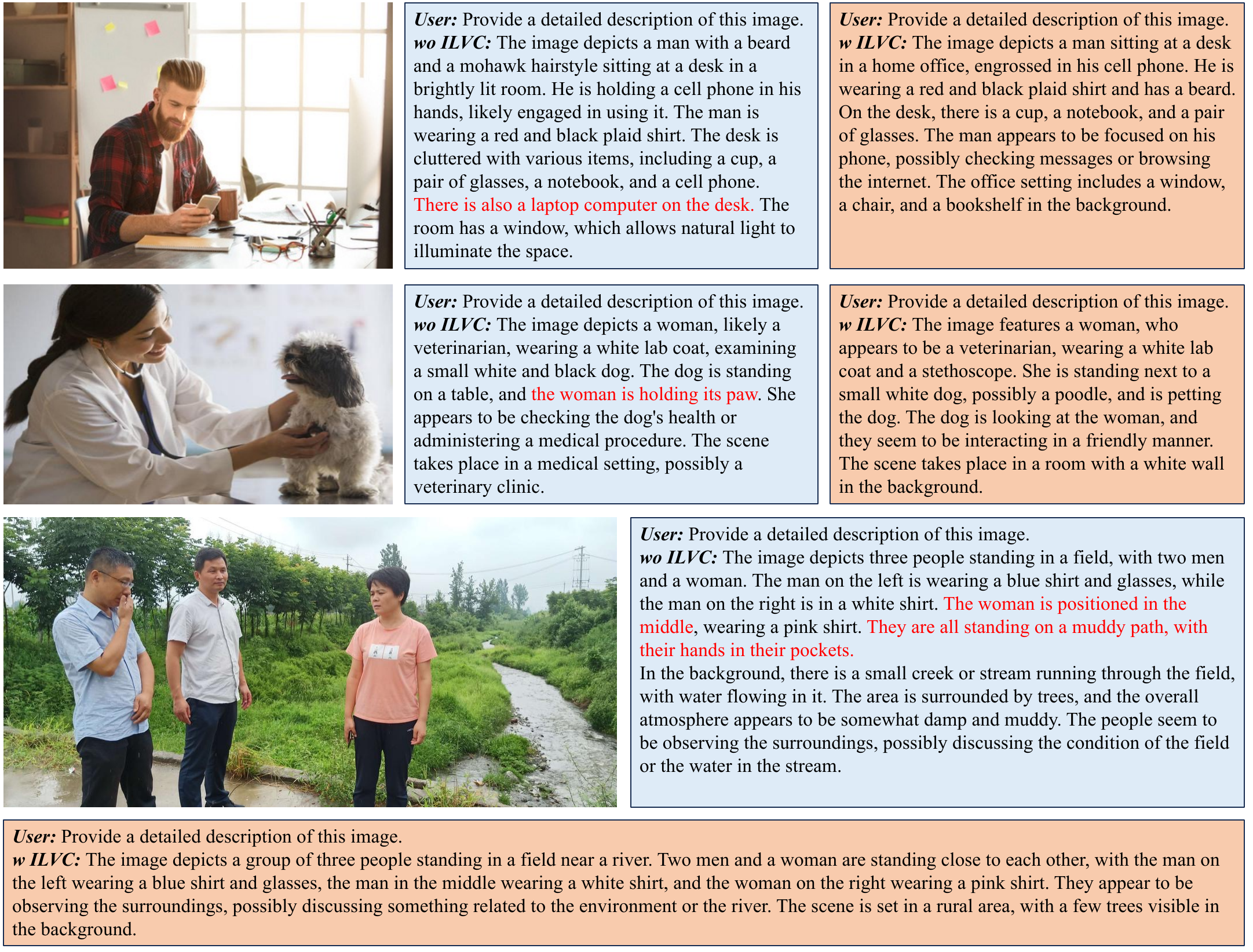}
       \caption{Comparison of Hallucinations in Image Caption \textit{wo/w} ILVC. The illusion content is marked in red.}
       \label{fig:hal}
    \end{figure*}

In Fig.~\ref{fig:hal}, we present additional visualization results to demonstrate the impact of employing ILVC on mitigating hallucination. As shown in the first sub-figure of Fig.\ref{fig:hal}, without ILVC, the model inaccurately generates the description, ``There is also a laptop computer on the desk." However, with ILVC applied, the model provides an accurate description without hallucinations. Similarly, in the second sub-figure, the model incorrectly infers the relationship between two objects, stating ``the woman is holding its paw" in the absence of ILVC. In the final sub-figure, without ILVC, the model suffers from significant hallucination, describing, ``They are all standing on a muddy path, with their hands in their pockets" a scenario that does not appear in the image. In contrast, with ILVC, the model produces a precise description, demonstrating the effectiveness of the ILVC strategy in reducing hallucinations.

    \begin{figure*}[t]
      \centering
       \includegraphics[width=0.90\linewidth]{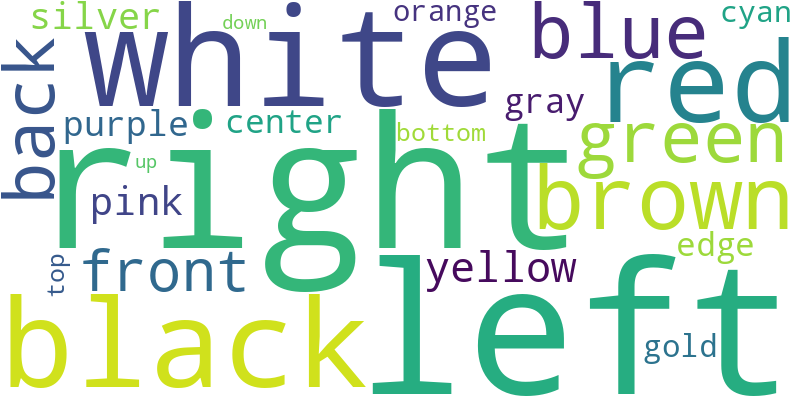}
       \caption{Word Cloud of the AttriEval Dataset.}
       \label{fig:wordcloud}
    \end{figure*}

    \begin{figure*}[t]
      \centering
       \includegraphics[width=0.95\linewidth]{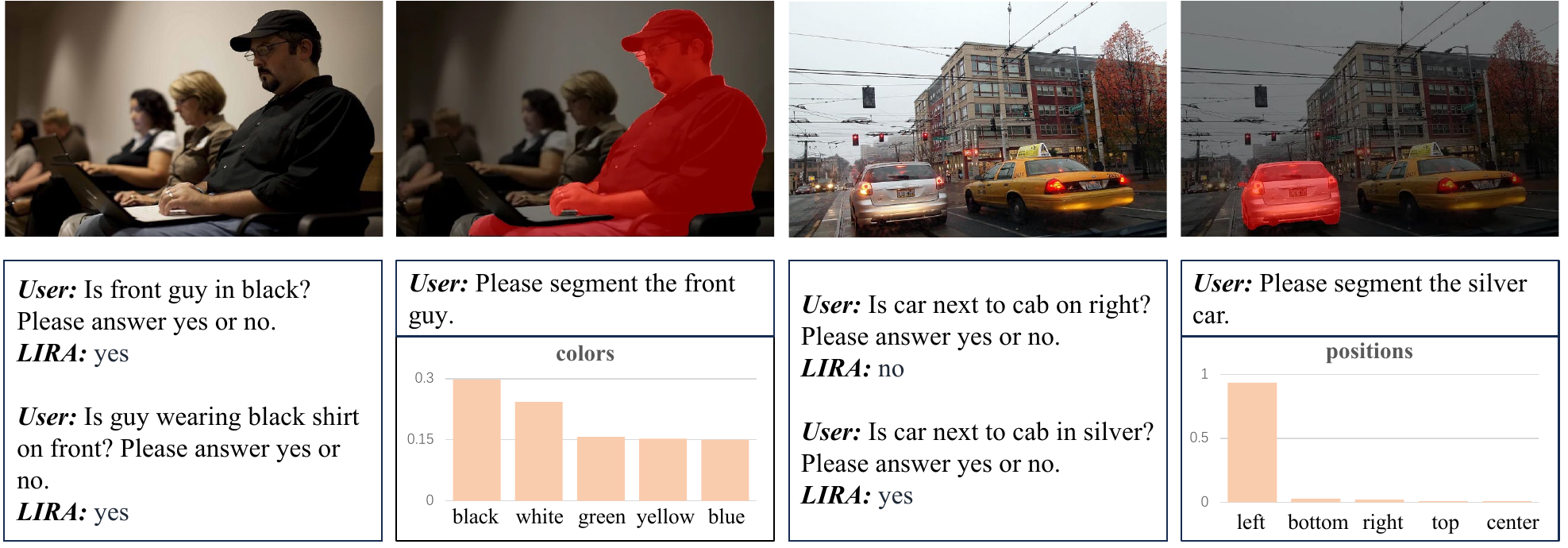}
       \caption{Visualization Results of LIRA on the AttriEval Dataset. The bar chart presents the top five attribute names with the highest probabilities for color or location.}
       \label{fig:attribute}
    \end{figure*}

\section{Details of the AttrEval Dataset}

AttrEval is a dataset specifically designed to evaluate the model's ability to understand object attributes. As shown in Fig.~\ref{fig:attribute}, it includes two types of tasks: Visual Question Answering (VQA) and Reference Segmentation (RefSeg), with 1436 and 618 samples, respectively. In the VQA task, the model needs to judge the attributes of objects. In the RefSeg task, the model must infer the object's attributes based on the logits corresponding to \texttt{<seg>} in the output.  We choose the RefCOCO dataset as the basis for constructing AttrEval. The process of building the dataset is as follows: We predefined a set of attribute categories, including category, location, and color. From multiple descriptions of the same object in RefCOCO, we extract unique attributes of color, location, and category. The specific attribute word cloud is shown in Fig.~\ref{fig:wordcloud}. Using these extracted attributes, we construct the VQA and RefSeg tasks based on different descriptions of the same object. For example, if one description of an object does not include color information, we use this description to refer to the object and ask a question about its color, requiring the model to predict the color attribute in the RefSeg task based on the logits corresponding to \texttt{<seg>} in the output. Similarly, if a description lacks location information, we ask a corresponding question about the object's location. In addition, while previous datasets, such as POPE, primarily focus on the existence of objects, our workplaces greater emphasis on the illusion of object attributes. 

As shown in Fig.~\ref{fig:attribute}, we ask questions about attributes that are not included in the description. For example, in the first image, the description ``front guy" does not contain the color attribute, so we ask the question, ``Is the front guy in black? Please answer yes or no." with the ground truth (GT) being ``yes". Additionally, for the question ``Is the front guy in black? Please answer yes or no." we also construct the question ``Is the front guy in white? Please answer yes or no." with the GT being ``no". Only when both of these questions are answered correctly do we consider the model to have correctly understood the color attribute of the ``front guy". In addition, Fig.~\ref{fig:attribute} also shows the visualization of LIRA's answers. LIRA correctly identifies the color and location attributes of the objects. Furthermore, in the RefSeg task, the logits corresponding to the \texttt{<seg>} token correctly contained the color or location attributes of the segmented object. For example, in the second image, LIRA correctly identifies the location of the ``silver car" as ``left" through the logits.

\newpage
\section{LIRA with Different Backbones}
To further validate the effectiveness of LIRA, we conduct experiments using Qwen2-1.5B~\cite{qwen2} from Qwen2VL~\cite{qwen2-vl}. As shown in Table~\ref{tab:qwen}, on the RefSeg task, LIRA-Qwen2-1.5B achieves an average score of 76.7\%, outperforming LIRA-InternLM2-1.8B by 1.2\%. On the comprehension task, it attains an average accuracy of 75.5\%. LIRA demonstrates strong performance with various backbones, achieving promising results on both the comprehension and segmentation tasks, thereby confirming its effectiveness and generalizability across different backbones.

\begin{table*}[]
    \centering  
    \scalebox{0.70}{
    \begin{tabular}{c|cccccccc|cccccccc}
    \toprule
                       &                          &                       &                         &                         &                         &                        &                             &                             & \multicolumn{3}{c}{RefCOCO} & \multicolumn{3}{c}{RefCOCO+} & \multicolumn{2}{c}{RefCOCOg} \\
    \multirow{-2}{*}{LLM} & \multirow{-2}{*}{VizWiz} & \multirow{-2}{*}{GQA} & \multirow{-2}{*}{VQAv2} & \multirow{-2}{*}{OKVQA} & \multirow{-2}{*}{SciQA} & \multirow{-2}{*}{POPE} & \multirow{-2}{*}{MMB-en} & \multirow{-2}{*}{MMB-cn} & Val     & TestA   & TestB   & Val     & TestA    & TestB   & Val           & Test         \\ \midrule

    InternLM2-1.8B            & 67.8                     & 61.1                  & 77.2                    & 53.7                    & 95.0                      & 89.1                  & 74.0                          & 71.7                        & 79.5    & 81.9    & 76.0      & 72.6    & 77.4     & 66.1    & 75.4          & 75.1
    \\
     InternLM2.5-7B            & 71.5                     & 63.5                  & 80.4                    & 62.9                    & 97.3                    & 88.1                   & 81.1                       & 80.5                       & 81.8    & 83.4    & 78.1    & 76.3    & 81.1     & 70.5    & 78.4          & 78.2         \\ 
    
     Qwen2-1.5B & 73.1                     & 62.5                  & 79.7                    & 58.5                    & 91.7                   & 87.3                   & 76.8                        &74.2                     & 80.8   & 82.3    & 77.1    & 74.6    & 78.2     & 68.3    & 76.7          & 75.9         \\   
    \bottomrule
    \end{tabular}}
    \caption{Performance with different LLMs.}
    \label{tab:qwen}
\end{table*}

\section{Risks of Error Accumulation and Instance Segmentation} 

ILVC may introduce error accumulation in multi-object segmentation, as inaccuracies in the initial masks can negatively impact the accuracy of subsequent segmentation results. However, we can use two different prompts to control whether to use ILVC to mitigate this. To investigate this, we follow PSALM and train on the COCO instance segmentation dataset, which features multi-object segmentation. When not using ILVC during training, the baseline mIou is 60.0. When using two prompts and 50\% of the data with ILVC and 50\% without during training, mIou is 60.6 when inference without ILVC, and mIou is 58.9 when inference with ILVC. The results demonstrate that using two prompts to control whether to use ILVC effectively reduces error accumulation, while incorporating ILVC during training improves instance segmentation performance by 0.6.

\section{Computational Overhead}
Although our method improves performance, it inevitably introduces some computational overhead, which remains within an acceptable range. The overall training time for LIRA-2B is approximately 22 hours, and the inference speed on referring segmentation tasks is around 21.6 tokens per second. Specifically, SEFE added 4 hours to the training time and reduced inference speed by 1.8 tokens per second. The ILVC module added 3 hours to training time, with no inference overhead for VQA tasks—since segmentation is not required—but resulted in a 1.3 tokens per second reduction for segmentation tasks.

\newpage
\section{Overall Pipeline}

To present our method’s workflow more clearly, we provide the following pseudocode.

\begin{algorithm}[H]
\caption{Overall Pipeline}
\label{alg:inference_simple}
\begin{algorithmic}[1]
\Require Global image $\mathbf{I}$, Text instruction $T_{ins}$
\Ensure Set of predicted masks $\mathcal{M}$, Generated output sequence $S_{out}$

\State $f_g, F_{pixel} \gets \text{SEFE}(\mathbf{I})$

\State $S \gets \{f_g, T_{ins}\}$
\State $\mathcal{M} \gets \emptyset$; \quad $M_{current} \gets \text{null}$

\While{not end-of-generation}
    \State $token \gets \text{LLM.generate}(S)$

    \If{$token$ is \texttt{<eos>}}
        \State \textbf{break}

    \ElsIf{$token$ is \texttt{<seg>}}
        \State $M_{current} \gets \text{PixelDecoder}(token, F_{pixel})$
        \State $\mathcal{M} \gets \mathcal{M} \cup \{M_{current}\}$
        \State $S \gets S \oplus token$
    \ElsIf{$token$ is \texttt{<image\_id>}}
        \State $I_l \gets \text{CropRegion}(\mathbf{I}, M_{current})$
        \State $f_l \gets \text{SemanticEncoder}(I_l)$
        \State $S \gets S \oplus token$
        \State $S \gets S \oplus f_l$
    \Else
        \State $S \gets S \oplus token$
    \EndIf
\EndWhile

\State $S_{out} \gets S$
\State \textbf{return} $\mathcal{M}, S_{out}$
\end{algorithmic}
\end{algorithm}

\end{document}